\documentclass{article}

\usepackage{arxiv}

\usepackage[utf8]{inputenc} 
\usepackage[T1]{fontenc}    
\usepackage{hyperref}       
\usepackage{url}            
\usepackage{booktabs}       
\usepackage{amsfonts}       
\usepackage{nicefrac}       
\usepackage{microtype}      
\usepackage{lipsum}		
\usepackage{graphicx}
\usepackage{natbib}
\usepackage{doi}

\title{Edge Devices Inference Performance Comparison}

\author{ 
{
Rafał Tobiasz}\thanks{Equal contribution.} \\
	Bulletprove sp. z o. o.\\
	Ignacego Mościckiego 1\\
    24-110 Puławy, Poland \\
	\texttt{rafal.tobiasz@bulletprove.com} \\
	\And
    {
    Grzegorz Wilczyński}\footnotemark[1] \\
	Bulletprove sp. z o. o.\\
	Ignacego Mościckiego 1\\
    24-110 Puławy, Poland \\
	\texttt{grzegorz.wilczynski@bulletprove.com} \\
 	\And
    {
    Piotr Graszka} \\
	Bulletprove sp. z o. o.\\
	Ignacego Mościckiego 1\\
    24-110 Puławy, Poland \\
	\texttt{piotr.graszka@bulletprove.com} \\
 	\And
    {
    Nikodem Czechowski} \\
	Bulletprove sp. z o. o.\\
	Ignacego Mościckiego 1\\
    24-110 Puławy, Poland \\
	\texttt{nikodem.czechowski@bulletprove.com} \\
 	\And
    {
    Sebastian Łuczak} \\
	Bulletprove sp. z o. o.\\
	Ignacego Mościckiego 1\\
    24-110 Puławy, Poland \\
	\texttt{sebastian.luczak@bulletprove.com} \\
}

\renewcommand{\shorttitle}{Edge Devices Inference Performance Comparison}

\hypersetup{
pdftitle={Edge Devices Inference Performance Comparison},
pdfsubject={cs.LG},
pdfauthor={Rafał Tobiasz et. al.},
pdfkeywords={Edge device, Deep learning, Computer vision},
}

\begin{document}
\maketitle

\begin{abstract}
In this work, we investigate the inference time of the MobileNet family, EfficientNet V1 and V2 family, VGG models, Resnet family, and InceptionV3 on four edge platforms. Specifically NVIDIA Jetson Nano, Intel Neural Stick, Google Coral USB Dongle, and Google Coral PCIe. Our main contribution is a thorough analysis of the aforementioned models in multiple settings, especially as a function of input size, the presence of the classification head, its size, and the scale of the model. Since throughout the industry, those architectures are mainly utilized as feature extractors we put our main focus on analyzing them as such. We show that Google platforms offer the fastest average inference time, especially for newer models like MobileNet or EfficientNet family, while Intel Neural Stick is the most universal accelerator allowing to run most architectures. These results should provide guidance for engineers in the early stages of AI edge systems development. All of them are accessible at \url{https://bulletprove.com/research/edge_inference_results.csv}.
\end{abstract}


\section{Introduction}
\label{s:intro}
A variety of applications exploit machine learning models, be it on a cloud \citep{gpt3, chatgpt, stable-diffusion}, personal computers, mobile devices \citep{blazeface, apple-face-detection}, or edge devices \citep{kws}. Especially for the latter we can observe intensified development of devices and suitable algorithms, since progressively more IoT applications use AI solutions.

A significant fraction of all those applications is in the computer vision domain, such as classification \citep{convnext, swin}, object detection \citep{yolov3, ssd}, image segmentation \citep{maskrcnn}. Those algorithms require vast amount of computational power to perform inference since the input data is high resolution.

Also, other reasons encourage the development of edge devices:
\begin{itemize}
    \item network load - sending high-resolution data from a vast amount of IoT devices to the computational unit may result in unwanted and unpredicted time delays \citep{edge-iot-survey, mobile-edge-survey},
    \item computational unit load - analyzing high-resolution data using current state-of-the-art models may result in a cost-inefficient system \citep{openai-compute},
    \item safety - sending raw data to the cloud may get targeted by hackers \citep{iot-security}\citep{iot-spy1, iot-spy2} or could lower the trust of a user who, e.g., for a face detection system, does not want his photos on an undisclosed server. Instead, it is better to use a feature extractor on an edge device and send only those anonymous features to the cloud.
\end{itemize}
To address those problems, many companies are targeting specialized inference chips, which result in a vast amount of different edge accelerators. Moreover, currently, various architectures are well-scalable and can extract features correctly. However, picking "the best" algorithm or platform is not possible. It depends on the application. Therefore, an engineer wanting to choose a platform and a model to start with, faces the time-consuming task of testing different variants. 

Few papers address model profiling on different AI accelerators, like in work done by \citet{inference-performance-paper}, although only for a few architectures (often with default parameters). Excellent research in this field is done by \citet{mlperf} which defines the proper way of different profiling methods. It allows users to perform their tests and share their results on a moderated platform.

However, none of those papers present any data regarding feature extractors. Authors analyze models only as classifiers. Whereas, those pre-trained models are mostly used as feature extractors. Therefore, analyzing them as classifiers may be slightly confusing in this setting. Profiling those algorithms only as feature extractors provides clear information on how long the first component of the final algorithm will last.

In this work, we present an extensive comparison of the most popular models available in TensorFlow \citep{tensorflow} Keras Applications: MobileNet family \citep{mobilenetv1, mobilenetv2, mobilenetv3}, EfficientNet (V1 and V2) family \citep{efficientnetv1, efficientnetv2}, ResNet (V1 and V2) family \citep{resnetv1, resnetv2}, VGG family \citep{vgg}, and InceptionV3 \citep{inceptionv3} on multiple platforms: NVIDIA Jetson Nano, Google Coral USB, Google Coral PCI, and Intel Neural Stick. Those algorithms were also analyzed in different settings, e.g., input sizes, scaling parameters, and more. Despite focusing on profiling models as feature extractors we also examined them with a classic ImageNet head (1000 classes) and a more real-life scenario (5 output neurons). 

The motivation behind this work is to create an in-depth comparison of the performance of different models on multiple edge devices so that it could make the work of fellow ML engineers more time- and cost-efficient.

\section{Edge AI Accelerators} \label{s:platforms}

AI on Edge is focusing on running artificial intelligence models on Edge devices \citep{ai-on-edge}. It favors low power consumption and small physical size at the cost of performance. An Edge AI accelerator is hardware specialized in processing AI workloads at the edge. Computation is local, close to data collection, which can be beneficial in preserving data privacy or in offline scenarios. Moreover, it reduces latency and communication costs when compared to Cloud AI.

\subsection{Google Coral Accelerator}

Google Coral Accelerator expands the user's system with an application-specific integrated circuit (ASIC) called Edge TPU, designed to deploy high-quality AI at the edge. Coral can perform 4 trillion operations per second using 2 watts of power. Edge TPU supports only 8-bit quantized Tensorflow Lite models compiled using a dedicated tool \citep{coral}. Potential use cases cover predictive maintenance, voice recognition, anomaly detection, machine vision, robotics, and more \citep{edge-tpu}. In this paper, we tested two IO Interface versions of Google Coral: PCIe and USB.

\subsection{Nvidia Jetson Nano}

Nvidia Jetson Nano is a small computer equipped with: 128-core NVIDIA Maxwell GPU, Quad-core ARM Cortex-A57 MPCore processor and 4GB of 64-bit LPDDR4 of memory. It is the only standalone edge device used in this comparison. The platform has an AI performance of 472 giga-floating point operations per second (GFLOPS) and uses 5 to 10 watts of power \citep{jetson-nano}. Jetson utilizes NVIDIA JetPack SDK, which provides useful features like CUDA, cuDNN and TensorRT. TensorRT optimizes inference of deep learning frameworks. For example, it allows for the use of FP16 precision for inference \citep{jetpack}. Potential use cases include predictive maintenance, voice recognition, anomaly detection, machine vision, robotics, patient monitoring, traffic management, and many more \citep{jetson-use-cases}.

\subsection{Intel Neural Compute Stick 2}

Neural Compute Stick 2 (NCS2) is a plug-and-play AI accelerator. It contains Intel Movidius Myriad X Vision Processing Unit (VPU), which includes 16 SHAVE cores and a dedicated DNN hardware accelerator. Intel Distribution of the OpenVINO toolkit is used to convert and optimize models for this platform. Potential use cases cover anomaly detection, machine vision, and more \citep{ncs2}.

\section{Methodology}

\subsection{Compared models}
TensorFlow is one of the most popular deep learning frameworks. Alongside many tools for creating, training, and profiling neural networks it also provides a set of already trained popular image classification networks. Those architectures are broadly used in the industry. We picked multiple algorithm families based on multiple motives.
\subsubsection{MobileNet}
This family of models consists of MobileNetV1, MobileNetV2, MobileNetV3-Small, and MobileNetV3-Large. They were particularly designed to run efficiently on mobile devices, hence the name. 

MobileNetV1 \citep{mobilenetv1} is based on a streamlined architecture that uses depthwise separable convolutions. The authors introduced two hyperparameters that efficiently tradeoff between latency and accuracy:
\begin{itemize}
    \item Width multiplier $\alpha$ - number of layer's input channels $M$ (where $M$ is the default number of channels) becomes $\alpha * M$, and the number of output channels $N$ becomes $\alpha * N$. It takes place for every layer. 
    \item Resolution multiplier $\rho$ - is implicitly set by setting the input resolution.
\end{itemize}
\noindent
Its fast inference is an effect of putting nearly all of the computation into dense 1x1 convolutions. It is crucial because they are implemented with highly optimized general matrix multiply (GEMM) functions and do not need any initial reordering in memory. 

MobileNetV2 \citep{mobilenetv2} is based on an inverted residual structure where the shortcut connections are between the thin bottleneck layers. Lightweight depthwise convolutions are here the source of non-linearity since they were removed in the narrow layers. The two prior introduced hyperparameters remain the same. This network improved the state-of-the-art for a wide range of performance points at that time for ImageNet \citep{imagenet}.

MobileNetV3 \citep{mobilenetv3} models leverage complementary search techniques complemented by the NetAdapt algorithm. The authors also introduced the swish activation function and rebuilt MobileNetV2's bottleneck by adding squeeze and excitation in the residual layer. MobileNetV3 family outperformed, at that time, the current state-of-the-art models on ImageNet taking into account top-1 accuracy and latency. 

\subsubsection{EfficientNet (V1 and V2)}
In their work \citep{efficientnetv1}, the authors improved model scaling and identified that balancing network depth, width, and resolution leads to better performance. They focused on optimizing FLOPS rather than latency since they did not target specific hardware. By using NAS (neural architecture search) they came up with a new scalable baseline network and called this family of models EfficientNets (now EfficientNets V1). There are eight models available in Tensorflow applications from B0 to B7.

In the following paper, the authors once again used a combination of NAS and scaling to optimize training speed and parameter efficiency. It resulted in the EfficientNetV2 \citep{efficientnetv2} family. It achieved the state-of-the-art top-1 accuracy on ImageNet, outperforming even the famous ViT \citep{vit}. This family consists of seven models, from B0 to B3 and S, M, and L. 

\subsubsection{InceptionV3}
As the Inception family of models had a crucial role in the development of convolutional neural networks for vision tasks, we decided to profile InceptionV3 \citep{inceptionv3}. Here authors aimed to utilize added computation more efficiently by factorizing convolutions and adding aggressive regularization. InceptionV3 achieved, at the time, the state-of-the-art top-1 error on ImageNet.

\subsubsection{VGG}
As with the former model, we decided to profile this \citep{vgg} family of networks based on historical reasons. The authors' main contribution was to increase the depth of the convolutional network using small 3x3 kernels. This allowed them to build 16-19 (VGG16, VGG19) layers architectures that earned first and second place at the ImageNet Challenge 2014 in the localization and classification tasks respectively. 

\subsubsection{ResNet (V1 and V2)}
In their work \citep{resnetv1}, the authors introduced residual connections that allowed training substantially deeper networks than their predecessors. On ImageNet they evaluated ResNet152 - networks 8x deeper than VGGs and yet having lower complexity - achieving state-of-the-art. We profiled ResNet50, ResNet101, and ResNet152 (we refer to this family as ResNetV1).

In the following work \citep{resnetv2}, the authors took on the propagation formulations behind the residual building blocks, trying to make the forward and backward signals flow easier. They rethought the residual blocks, moreover removed ReLU from the "easiest" path after the addition operation. For the ResNetV2 family, we profiled analogous models to its predecessor. We profile ResNet families because of the same reasons as InceptionV3 and VGGs.

\subsection{Model parameters}
The only measurement performed in this work is the inference time. As we wanted to profile the influence of different model hyperparameters, we created extensive sets of architectures for every model family. Each neural network is built with different parameters.

\subsubsection{Input size}
Nowadays, AI applications need to analyze images of various sizes, ranging from $224^{2}$ to $1024^{2}$ or larger. This trend is also noticeable on edge devices. Taking into account restricted computational resources an engineer has to be able to pick a suitable architecture that will allow using the model efficiently, e.g. in real-time. 

\subsubsection{Preprocessing}
Every analyzed model has some specific input preprocessing, e.g., reducing the mean and dividing by standard deviation. It is important to know how such simple data alteration will affect the model's inference time.

\subsubsection{Classification heads}
In this work, our key focus is on analyzing models as feature extractors (no classification heads). Currently, those architectures are used in this fashion in the industry hence it is crucial to know how computing features will affect the inference time. Furthermore, we profiled architectures with classic ImageNet head (1000 classes) to make our results comparable to others. Last but not least, we analyzed smaller classification heads (5 classes) which represent more "real-life" case scenarios for image classification.

\subsubsection{Width multiplier $\alpha$}
The aforementioned parameter is specific only to the MobileNet family which allows efficient scaling of the models.

\subsubsection{Precision}
Only applicable to models on Jetson Nano, possible types are FLOAT32 and FLOAT16 (this board does not support INT8).

\subsection{Benchmark prerequisites}
Each platform imposes different model preparation techniques as well as environment requirements. 

\subsubsection{Google Coral}
Each model has to be quantized, converted to a TFLite format, and compiled with the edgetpu-compiler.
\bigbreak
\noindent
The Coral USB dongle was tested on Ubuntu 20.04.5 LTS with Intel Core i5-1135G7 2.4GHz,  32GB of RAM, a USB 3.0 port, and Edge TPU runtime for Linux that operates at the maximum clock frequency \citep{coral}.
\bigbreak
\noindent
The Coral PCIe was tested on Radxa CM3 IO Board with Ubuntu 20.04.4 LTS, 4GB of RAM, and, analogous to PC, Edge TPU runtime.
\bigbreak
\noindent
The coral benchmarking script is based on \citep{pycoral-repo}, which is also the reason for using \textit{timeit} \citep{timeit} package. 
\subsubsection{Inter Neural Stick}
Models were not quantized, as Neural Stick does not support quantization, only converted to ONNX format with \textit{tf2onnx} package \citep{tf2onnx}. 
\bigbreak
\noindent
The Intel Neural Stick was tested on Ubuntu 20.04.4 LTS with AMD Ryzen 5 1600, 128GB of RAM, and a USB 3.0 port. Moreover, it needed OpenVino-dev with additional libraries (like TensorFlow, and ONNX) \citep{openvino-dev}, an OpenVino toolkit \citep{openvino-package}, and set up variables with \textit{setupvars.sh} from the prior downloaded toolkit. 
\bigbreak
\noindent
The Neural Stick benchmarking script is based on the \textit{hello\_classification.py} script from the OpenVino toolkit \citep{openvino-package}. Additionally, models were compiled with the latency hint. 
\subsubsection{Nvidia Jetson Nano}
Each model is saved to Protocol Buffers format, and then, is converted to TensorRT \citep{tensorrt} architecture using TensorFlow's experimental converter. To make experiments quicker we compiled TensorRT models on RTX3090 and then built them on the target device.

\subsection{Benchmarking process}
Inferences in our tests are synchronous and blocking, similar to the single-stream scenario described by Reddi \citep{mlperf}, i.e., the batch size is equal to 1, but our metric is the whole latency time. 
The Benchmarking script is written in Python programming language, and inference time is measured with the \textit{timeit} package. Models are analyzed in a grid search fashion over all possible parameters: $input\_size$, $preprocessing$, $classification\_head$, $\alpha$, and $precision$. The process runs as follows:
\begin{itemize}
    \item model creation from TensorFlow's application module,
    \item outcome compilation to meet platform-specific requirements,
    \item an input image creation of size $[input\_size, input\_size, 3]$ with platform-specific data type,
    \item warmup inferences with a compiled model on a selected platform (10 for every run),
    \item proper inferences along with time measurement.
\end{itemize}

\begin{figure*}[!htbp]
	\centering
    \includegraphics[width=0.85\textwidth]{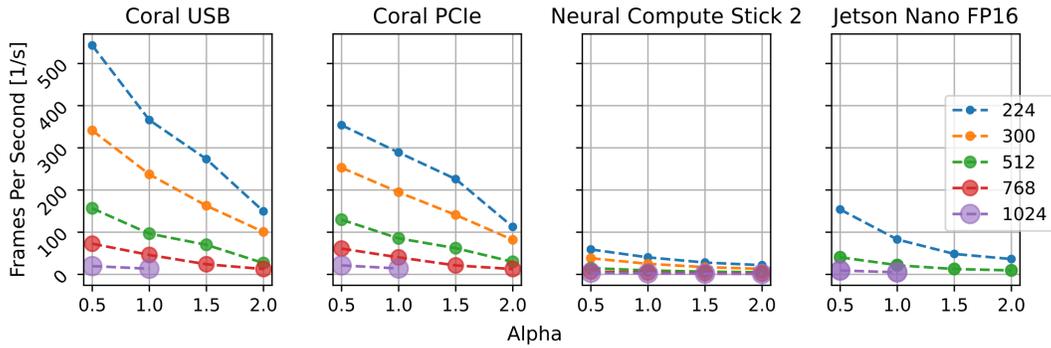}
	\caption{
Scaling up MobileNetV2 alpha parameter on four platforms. Missing data points are the result of compilation errors.
}
	\label{fig:mobilenetv2}
\end{figure*}

\begin{figure*}[!htbp]
	\centering
    \includegraphics[width=0.85\textwidth]{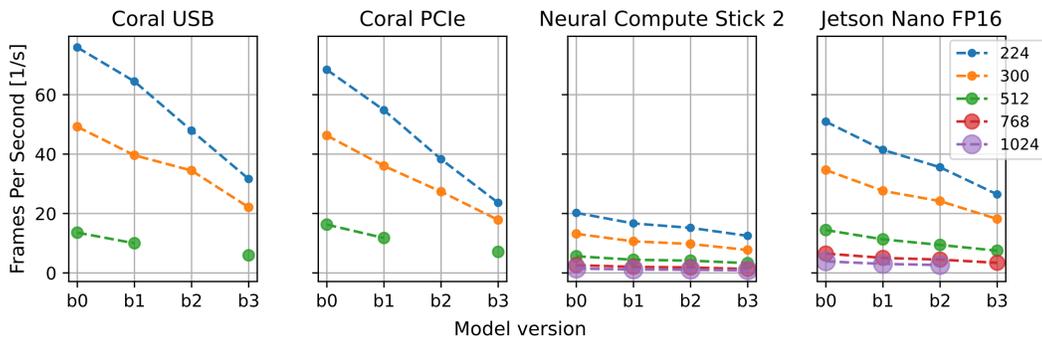}
	\caption{
Scaling EfficientNetV2 on four platforms. Missing data points on EfficientNetV2B2 are the result of compilation errors.
}
	\label{fig:efficientnetv2}
\end{figure*}

\begin{figure*}[!htbp]
	\centering
    \includegraphics[width=0.85\textwidth]{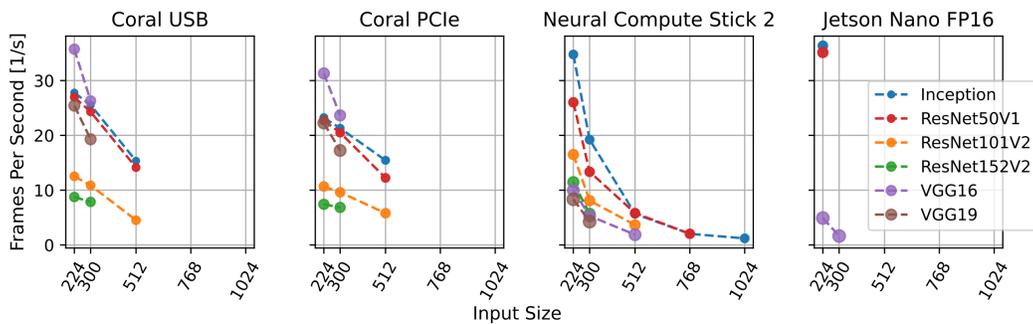}
	\caption{
Model input size vs frames vs platform on larger model families. Missing data points for specific networks are mainly the result of exceeding the platform's model size limits.
}
	\label{fig:others}
\end{figure*}

At the end of every grid search run, the script creates a CSV file with
\begin{itemize}
    \item model parameters,
    \item minimum inference time,
    \item maximum inference time,
    \item mean inference time,
    \item the standard deviation of inference time,
    \item median inference time for Jetson Nano (skewness of inference time distribution).
\end{itemize}

The default number of proper inferences is 1024. In special cases, it might vary, e.g., for the MobileNetV3 family on the Neural Stick it was 128 caused by the duration of the inference.

\section{Results}

Our experiment covered 3095 unique test cases, focusing predominantly on MobileNet and EfficientNet model families. For visualizations and Table \ref{t:results} results, we used models which include input data preprocessing, and have no classification heads. Those are the results that we wanted to focus on, since, in this work, we first analyze architectures for being time-efficient feature extractors. Those picked results provide good insight into the rest of the collected data.

All test results are available in a CSV format at \url{https://bulletprove.com/research/edge_inference_results.csv}. 

\subsection{Performance}

Figure \ref{fig:mobilenetv2} reveals the platform's performance on MobileNetV2. Coral devices achieve the highest frames-per-second performance. NCS2 was the only device to successfully run models with all input sizes, which makes this platform the most versatile.

Figure \ref{fig:efficientnetv2} shows that the Coral is generally faster than NCS2 and Jetson Nano in this model group, especially in the case of smaller input size. Apart from NCS2, platforms had difficulties running models with larger input sizes.

Figure \ref{fig:others} reflects the inference results on larger models. For models with input size 224 and outside the VGG family, the NCS2 is the fastest. Moreover, the device covers the broadest range of model configurations. With increasing model input size Coral slightly outperforms other platforms. Due to memory errors, only four inference test cases were completed successfully for Jetson Nano.

What also can be seen in figures \ref{fig:mobilenetv2}, \ref{fig:efficientnetv2}, \ref{fig:others} is the difference in the slope of curves, which indicates that inference time scales differently with the number of parameters across compared platforms. 

As shown in Table \ref{t:results}, for almost every unique architecture and input size setting Coral (either USB Dongle or PCIe) is the best platform. The only exception occurs for the InceptionV3 and Resnet50 models however, it is crucial to mention that these architectures do not fit entirely on Google's platform. Nevertheless, the performance gap between Google's platforms and the others deepens with increasing the input size, which indicates better computation optimization for the prior platforms. 

Results for Jetson Nano have, on average, $8.77$ times bigger standard deviation of inference time than for other platforms. This value is based on data from Table \ref{t:results} and only for MobileNetV2 and EfficientNetV2B0 however, this trend is correct for all measurements. It is a valuable insight for, e.g., designing a system with a strict maximum inference time constraint. 

For MobileNetV2, architecture designed especially for mobile devices, we can observe a significant difference in FPS between Corals and other platforms. For an input size of 224 Coral USB was $4.42$ times faster than Jetson and $9.08$ than Neural Stick. In addition, Corals performed better for an input size of 512 than other devices for the smaller one - $1.16$ times faster than Jetson Nano and $2.39$ than Neural Stick.

\begin{table*}
\centering
\caption{\textbf{Inference time comparison for every model family representative per platform and input size.} Bolded rows show the best result for a certain model and input size, whereas those without measurements indicate an inability of running that setting. Units: Input size is in pixels, FPS (frames per second) - $1/s$, and the rest of the time variables are in $ms$.}
\label{t:results}
\begin{tabular}{cccccccc}
\toprule
    Platform &       Model name &  Input Size &    FPS &  Mean time &  Std time &  Min time &  Max time \\
\midrule
 \textbf{Coral USB} &      \textbf{MobileNetV2} &         \textbf{224} & \textbf{365.82} &       \textbf{2.73} &      \textbf{0.11} &      \textbf{2.41} &      \textbf{3.02} \\
  Coral PCIe &      MobileNetV2 &         224 & 289.01 &       3.46 &      0.24 &      2.55 &      4.69 \\
   \textbf{Coral USB} &      \textbf{MobileNetV2} &         \textbf{512} &  \textbf{96.51} &      \textbf{10.36} &      \textbf{0.08} &     \textbf{10.05} &     \textbf{10.57} \\
  Coral PCIe &      MobileNetV2 &         512 &  85.33 &      11.72 &      0.41 &     10.56 &     24.30 \\
 Jetson FP16 &      MobileNetV2 &         224 &  82.75 &      12.08 &      1.91 &      5.20 &     17.58 \\
   \textbf{Coral USB} & \textbf{EfficientNetV2B0} &         \textbf{224} &  \textbf{75.91} &      \textbf{13.17} &      \textbf{0.41} &     \textbf{11.77} &     \textbf{14.09} \\
  Coral PCIe & EfficientNetV2B0 &         224 &  68.39 &      14.62 &      0.15 &     12.88 &     15.03 \\
 Jetson FP16 & EfficientNetV2B0 &         224 &  50.92 &      19.64 &      1.11 &     11.54 &     24.46 \\
Neural Stick &      MobileNetV2 &         224 &  40.31 &      24.81 &      0.31 &     24.06 &     25.44 \\
\textbf{Jetson FP16} &      \textbf{InceptionV3} &         \textbf{224} &  \textbf{36.35} &      \textbf{27.51} &      \textbf{4.49} &      \textbf{6.75} &     \textbf{47.51} \\
   \textbf{Coral USB} &            \textbf{VGG16} &         \textbf{224} &  \textbf{35.73} &      \textbf{27.99} &      \textbf{0.13} &     \textbf{27.41} &     \textbf{28.30} \\
\textbf{Jetson FP16} &         \textbf{Resnet50} &         \textbf{224} &  \textbf{35.13} &      \textbf{28.47} &      \textbf{4.98} &      \textbf{4.47} &     \textbf{46.09} \\
Neural Stick &      InceptionV3 &         224 &  34.76 &      28.77 &      0.28 &     27.70 &     29.55 \\
  Coral PCIe &            VGG16 &         224 &  31.31 &      31.94 &      0.16 &     31.04 &     34.92 \\
   Coral USB &      InceptionV3 &         224 &  27.78 &      35.99 &      0.23 &     35.09 &     36.45 \\
   Coral USB &         Resnet50 &         224 &  26.96 &      37.09 &      0.18 &     36.07 &     37.48 \\
Neural Stick &         Resnet50 &         224 &  26.03 &      38.42 &      0.27 &     37.59 &     39.14 \\
  Coral PCIe &      InceptionV3 &         224 &  23.23 &      43.05 &      0.14 &     41.92 &     43.99 \\
  Coral PCIe &         Resnet50 &         224 &  22.71 &      44.04 &      0.14 &     43.09 &     44.68 \\
 Jetson FP16 &      MobileNetV2 &         512 &  21.91 &      45.64 &      8.35 &      6.82 &     84.63 \\
Neural Stick & EfficientNetV2B0 &         224 &  20.20 &      49.50 &      0.38 &     48.18 &     50.26 \\
  \textbf{Coral PCIe} & \textbf{EfficientNetV2B0} &         \textbf{512} &  \textbf{16.26} &      \textbf{61.50} &      \textbf{0.21} &     \textbf{58.74} &     \textbf{62.63} \\
  \textbf{Coral PCIe} &      \textbf{InceptionV3} &         \textbf{512} &  \textbf{15.44} &      \textbf{64.75} &      \textbf{0.14} &     \textbf{63.61} &     \textbf{65.83} \\
   Coral USB &      InceptionV3 &         512 &  15.35 &      65.15 &      1.54 &     62.04 &     69.33 \\
 Jetson FP16 & EfficientNetV2B0 &         512 &  14.43 &      69.30 &      5.50 &     41.00 &     95.82 \\
   \textbf{Coral USB} &         \textbf{Resnet50} &         \textbf{512} &  \textbf{14.13} &      \textbf{70.75} &      \textbf{0.22} &     \textbf{69.59} &     \textbf{71.09} \\
   Coral USB & EfficientNetV2B0 &         512 &  13.56 &      73.76 &      2.53 &     70.05 &     83.77 \\
  Coral PCIe &         Resnet50 &         512 &  12.24 &      81.71 &      0.19 &     79.10 &     83.16 \\
Neural Stick &            VGG16 &         224 &   9.99 &     100.10 &      0.27 &     99.34 &    100.98 \\
Neural Stick &      MobileNetV2 &         512 &   9.35 &     106.94 &      0.48 &    105.60 &    108.00 \\
Neural Stick &         Resnet50 &         512 &   5.82 &     171.78 &      0.66 &    169.68 &    175.44 \\
Neural Stick &      InceptionV3 &         512 &   5.68 &     176.08 &      0.62 &    173.58 &    177.87 \\
Neural Stick & EfficientNetV2B0 &         512 &   5.62 &     178.09 &      0.46 &    176.53 &    179.37 \\
 Jetson FP16 &            VGG16 &         224 &   4.92 &     203.45 &     42.62 &      5.64 &    332.42 \\
\textbf{Neural Stick} &            \textbf{VGG16} &         \textbf{512} &   \textbf{1.87} &     \textbf{533.86} &      \textbf{0.71} &    \textbf{531.54} &    \textbf{536.14} \\
  Coral PCIe &            VGG16 &         512 &      - &          - &         - &         - &         - \\
   Coral USB &            VGG16 &         512 &      - &          - &         - &         - &         - \\
 Jetson FP16 &            VGG16 &         512 &      - &          - &         - &         - &         - \\
 Jetson FP16 &      InceptionV3 &         512 &      - &          - &         - &         - &         - \\
 Jetson FP16 &         Resnet50 &         512 &      - &          - &         - &         - &         - \\
\bottomrule
\end{tabular}
\end{table*}

\subsection{Limitations}

On Google Coral, when the model size exceeds on-chip memory size limits, the model's data has to be fetched from the external memory, which results in additional latency. Exceeding the unspecified model size limit on Google Coral results in a compilation failure. 

Coral does not support the hard-swish activation function, which is required to compile MobileNetV3 directly in a way that allows full TPU computation. Therefore, some operations are executed off-chip, increasing model latency.

For large enough models, Neural Stick does not behave like Google's platform (using on-chip and off-chip memory), i.e. throws $NC\_OUT\_OF\_MEMORY$ error terminating running script. However, it is still able to work with more models than Coral. 

In the case of Jetson Nano, it takes significantly longer to prepare an inference model compared to other platforms. It turned out to be a bottleneck of our experiment. Similarly to Neural Stick, exceeding GPU's RAM results in an out-of-memory error.

\section{Conclusions}

This paper presents an extensive inference time performance comparison on Edge AI devices, specifically: NVIDIA Jetson Nano, Google Coral USB, Google Coral PCI, and Intel Neural Stick. For inference, we use variations of model families: MobileNet, EfficientNet, ResNet, VGG, and InceptionV3. Test configurations included mainly changes in the model's input size, classification head, type and scale. The experiments' results indicate that Google Coral is the platform that offers the fastest average inference time. Jetson Nano inference tests suggest that the platform is prone to latency spikes, which is undesirable in time-restricted use cases. The NCS2 platform is the most universal considering the model type and model size choice in the scope of this experiment. We hope this benchmark will help engineers in developing AI at the edge.

\section{Acknowledgements}
N.C. and P.G. conceived the experiment, R.T. and G.W. carried out the measurements and prepared the figures under N.C. supervision. G.W. and R.T. wrote the manuscript. S.L. supervised the project. All authors discussed the results and contributed to the final manuscript.

This research was undertaken as part of the “BulletProve - a system for supporting the training of long-distance shooters using machine learning methods” project and is jointly funded by the National Centre for Research and Development (Szybka Ścieżka 0552/21).

\bibliographystyle{unsrtnat}
\bibliography{ms}  

\begin{thebibliography}{44}
\providecommand{\natexlab}[1]{#1}
\providecommand{\url}[1]{\texttt{#1}}
\expandafter\ifx\csname urlstyle\endcsname\relax
  \providecommand{\doi}[1]{doi: #1}\else
  \providecommand{\doi}{doi: \begingroup \urlstyle{rm}\Url}\fi

\bibitem[Brown et~al.(2020)Brown, Mann, Ryder, Subbiah, Kaplan, Dhariwal,
  Neelakantan, Shyam, Sastry, Askell, Agarwal, Herbert{-}Voss, Krueger,
  Henighan, Child, Ramesh, Ziegler, Wu, Winter, Hesse, Chen, Sigler, Litwin,
  Gray, Chess, Clark, Berner, McCandlish, Radford, Sutskever, and Amodei]{gpt3}
Tom~B. Brown, Benjamin Mann, Nick Ryder, Melanie Subbiah, Jared Kaplan,
  Prafulla Dhariwal, Arvind Neelakantan, Pranav Shyam, Girish Sastry, Amanda
  Askell, Sandhini Agarwal, Ariel Herbert{-}Voss, Gretchen Krueger, Tom
  Henighan, Rewon Child, Aditya Ramesh, Daniel~M. Ziegler, Jeffrey Wu, Clemens
  Winter, Christopher Hesse, Mark Chen, Eric Sigler, Mateusz Litwin, Scott
  Gray, Benjamin Chess, Jack Clark, Christopher Berner, Sam McCandlish, Alec
  Radford, Ilya Sutskever, and Dario Amodei.
\newblock Language models are few-shot learners.
\newblock \emph{CoRR}, abs/2005.14165, 2020.
\newblock URL \url{https://arxiv.org/abs/2005.14165}.

\bibitem[OpenAI(2022)]{chatgpt}
OpenAI.
\newblock Chatgpt.
\newblock \url{https://openai.com/blog/chatgpt/}, 2022.

\bibitem[Rombach et~al.(2021)Rombach, Blattmann, Lorenz, Esser, and
  Ommer]{stable-diffusion}
Robin Rombach, Andreas Blattmann, Dominik Lorenz, Patrick Esser, and
  Bj{\"{o}}rn Ommer.
\newblock High-resolution image synthesis with latent diffusion models.
\newblock \emph{CoRR}, abs/2112.10752, 2021.
\newblock URL \url{https://arxiv.org/abs/2112.10752}.

\bibitem[Bazarevsky et~al.(2019)Bazarevsky, Kartynnik, Vakunov, Raveendran, and
  Grundmann]{blazeface}
Valentin Bazarevsky, Yury Kartynnik, Andrey Vakunov, Karthik Raveendran, and
  Matthias Grundmann.
\newblock Blazeface: Sub-millisecond neural face detection on mobile gpus.
\newblock \emph{CoRR}, abs/1907.05047, 2019.
\newblock URL \url{http://arxiv.org/abs/1907.05047}.

\bibitem[Team(2017)]{apple-face-detection}
Apple~ML Team.
\newblock Apple face detection on mobile.
\newblock \url{https://machinelearning.apple.com/research/face-detection},
  2017.

\bibitem[Zhang et~al.(2017)Zhang, Suda, Lai, and Chandra]{kws}
Yundong Zhang, Naveen Suda, Liangzhen Lai, and Vikas Chandra.
\newblock Hello edge: Keyword spotting on microcontrollers.
\newblock \emph{CoRR}, abs/1711.07128, 2017.
\newblock URL \url{http://arxiv.org/abs/1711.07128}.

\bibitem[Liu et~al.(2022)Liu, Mao, Wu, Feichtenhofer, Darrell, and
  Xie]{convnext}
Zhuang Liu, Hanzi Mao, Chao{-}Yuan Wu, Christoph Feichtenhofer, Trevor Darrell,
  and Saining Xie.
\newblock A convnet for the 2020s.
\newblock \emph{CoRR}, abs/2201.03545, 2022.
\newblock URL \url{https://arxiv.org/abs/2201.03545}.

\bibitem[Liu et~al.(2021)Liu, Lin, Cao, Hu, Wei, Zhang, Lin, and Guo]{swin}
Ze~Liu, Yutong Lin, Yue Cao, Han Hu, Yixuan Wei, Zheng Zhang, Stephen Lin, and
  Baining Guo.
\newblock Swin transformer: Hierarchical vision transformer using shifted
  windows.
\newblock \emph{CoRR}, abs/2103.14030, 2021.
\newblock URL \url{https://arxiv.org/abs/2103.14030}.

\bibitem[Redmon and Farhadi(2018)]{yolov3}
Joseph Redmon and Ali Farhadi.
\newblock Yolov3: An incremental improvement.
\newblock \emph{CoRR}, abs/1804.02767, 2018.
\newblock URL \url{http://arxiv.org/abs/1804.02767}.

\bibitem[Liu et~al.(2015)Liu, Anguelov, Erhan, Szegedy, Reed, Fu, and
  Berg]{ssd}
Wei Liu, Dragomir Anguelov, Dumitru Erhan, Christian Szegedy, Scott~E. Reed,
  Cheng{-}Yang Fu, and Alexander~C. Berg.
\newblock {SSD:} single shot multibox detector.
\newblock \emph{CoRR}, abs/1512.02325, 2015.
\newblock URL \url{http://arxiv.org/abs/1512.02325}.

\bibitem[He et~al.(2017)He, Gkioxari, Doll{\'{a}}r, and Girshick]{maskrcnn}
Kaiming He, Georgia Gkioxari, Piotr Doll{\'{a}}r, and Ross~B. Girshick.
\newblock Mask {R-CNN}.
\newblock \emph{CoRR}, abs/1703.06870, 2017.
\newblock URL \url{http://arxiv.org/abs/1703.06870}.

\bibitem[Yu et~al.(2018)Yu, Liang, He, Hatcher, Lu, Lin, and
  Yang]{edge-iot-survey}
Wei Yu, Fan Liang, Xiaofei He, William~Grant Hatcher, Chao Lu, Jie Lin, and
  Xinyu Yang.
\newblock A survey on the edge computing for the internet of things.
\newblock \emph{IEEE Access}, 6:\penalty0 6900--6919, 2018.
\newblock \doi{10.1109/ACCESS.2017.2778504}.

\bibitem[Mao et~al.(2017)Mao, You, Zhang, Huang, and
  Letaief]{mobile-edge-survey}
Yuyi Mao, Changsheng You, Jun Zhang, Kaibin Huang, and Khaled~B. Letaief.
\newblock A survey on mobile edge computing: The communication perspective.
\newblock \emph{IEEE Communications Surveys and Tutorials}, 19\penalty0
  (4):\penalty0 2322--2358, 2017.
\newblock \doi{10.1109/COMST.2017.2745201}.

\bibitem[Amodei and Hernandez(2018)]{openai-compute}
Dario Amodei and Danny Hernandez.
\newblock Ai and compute.
\newblock \url{https://openai.com/blog/ai-and-compute/}, 2018.

\bibitem[Neshenko et~al.(2019)Neshenko, Bou-Harb, Crichigno, Kaddoum, and
  Ghani]{iot-security}
Nataliia Neshenko, Elias Bou-Harb, Jorge Crichigno, Georges Kaddoum, and Nasir
  Ghani.
\newblock Demystifying iot security: An exhaustive survey on iot
  vulnerabilities and a first empirical look on internet-scale iot
  exploitations.
\newblock \emph{IEEE Communications Surveys and Tutorials}, 21\penalty0
  (3):\penalty0 2702--2733, 2019.
\newblock \doi{10.1109/COMST.2019.2910750}.

\bibitem[Franceschi-Bicchierai(2017{\natexlab{a}})]{iot-spy1}
Lorenzo Franceschi-Bicchierai.
\newblock How this internet of things stuffed animal can be remotely turned
  into a spy device.
\newblock
  \url{https://www.vice.com/en/article/qkm48b/how-this-internet-of-things-teddy-bear-can-be-remotely-turned-into-a-spy-device},
  2017{\natexlab{a}}.

\bibitem[Franceschi-Bicchierai(2017{\natexlab{b}})]{iot-spy2}
Lorenzo Franceschi-Bicchierai.
\newblock Internet of things teddy bear leaked 2 million parent and kids
  message recordings.
\newblock
  \url{https://www.vice.com/en/article/pgwean/internet-of-things-teddy-bear-leaked-2-million-parent-and-kids-message-recordings},
  2017{\natexlab{b}}.

\bibitem[Reza et~al.(2021)Reza, Yan, Dong, and
  Qian]{inference-performance-paper}
Sheikh~Rufsan Reza, Yuzhong Yan, Xishuang Dong, and Lijun Qian.
\newblock Inference performance comparison of convolutional neural networks on
  edge devices.
\newblock In Sara Paiva, S{\'e}rgio~Ivan Lopes, Rafik Zitouni, Nishu Gupta,
  S{\'e}rgio~F. Lopes, and Takuro Yonezawa, editors, \emph{Science and
  Technologies for Smart Cities}, pages 323--335, Cham, 2021. Springer
  International Publishing.
\newblock ISBN 978-3-030-76063-2.

\bibitem[Reddi et~al.(2019)Reddi, Cheng, Kanter, Mattson, Schmuelling, Wu,
  Anderson, Breughe, Charlebois, Chou, Chukka, Coleman, Davis, Deng, Diamos,
  Duke, Fick, Gardner, Hubara, Idgunji, Jablin, Jiao, John, Kanwar, Lee, Liao,
  Lokhmotov, Massa, Meng, Micikevicius, Osborne, Pekhimenko, Rajan, Sequeira,
  Sirasao, Sun, Tang, Thomson, Wei, Wu, Xu, Yamada, Yu, Yuan, Zhong, Zhang, and
  Zhou]{mlperf}
Vijay~Janapa Reddi, Christine Cheng, David Kanter, Peter Mattson, Guenther
  Schmuelling, Carole-Jean Wu, Brian Anderson, Maximilien Breughe, Mark
  Charlebois, William Chou, Ramesh Chukka, Cody Coleman, Sam Davis, Pan Deng,
  Greg Diamos, Jared Duke, Dave Fick, J.~Scott Gardner, Itay Hubara, Sachin
  Idgunji, Thomas~B. Jablin, Jeff Jiao, Tom~St. John, Pankaj Kanwar, David Lee,
  Jeffery Liao, Anton Lokhmotov, Francisco Massa, Peng Meng, Paulius
  Micikevicius, Colin Osborne, Gennady Pekhimenko, Arun Tejusve~Raghunath
  Rajan, Dilip Sequeira, Ashish Sirasao, Fei Sun, Hanlin Tang, Michael Thomson,
  Frank Wei, Ephrem Wu, Lingjie Xu, Koichi Yamada, Bing Yu, George Yuan, Aaron
  Zhong, Peizhao Zhang, and Yuchen Zhou.
\newblock Mlperf inference benchmark, 2019.

\bibitem[Google(2023{\natexlab{a}})]{tensorflow}
Google.
\newblock Tensorflow.
\newblock \url{https://www.tensorflow.org/}, 2023{\natexlab{a}}.

\bibitem[Howard et~al.(2017)Howard, Zhu, Chen, Kalenichenko, Wang, Weyand,
  Andreetto, and Adam]{mobilenetv1}
Andrew~G. Howard, Menglong Zhu, Bo~Chen, Dmitry Kalenichenko, Weijun Wang,
  Tobias Weyand, Marco Andreetto, and Hartwig Adam.
\newblock Mobilenets: Efficient convolutional neural networks for mobile vision
  applications.
\newblock \emph{CoRR}, abs/1704.04861, 2017.
\newblock URL \url{http://arxiv.org/abs/1704.04861}.

\bibitem[Sandler et~al.(2018)Sandler, Howard, Zhu, Zhmoginov, and
  Chen]{mobilenetv2}
Mark Sandler, Andrew~G. Howard, Menglong Zhu, Andrey Zhmoginov, and
  Liang{-}Chieh Chen.
\newblock Inverted residuals and linear bottlenecks: Mobile networks for
  classification, detection and segmentation.
\newblock \emph{CoRR}, abs/1801.04381, 2018.
\newblock URL \url{http://arxiv.org/abs/1801.04381}.

\bibitem[Howard et~al.(2019)Howard, Sandler, Chu, Chen, Chen, Tan, Wang, Zhu,
  Pang, Vasudevan, Le, and Adam]{mobilenetv3}
Andrew Howard, Mark Sandler, Grace Chu, Liang{-}Chieh Chen, Bo~Chen, Mingxing
  Tan, Weijun Wang, Yukun Zhu, Ruoming Pang, Vijay Vasudevan, Quoc~V. Le, and
  Hartwig Adam.
\newblock Searching for mobilenetv3.
\newblock \emph{CoRR}, abs/1905.02244, 2019.
\newblock URL \url{http://arxiv.org/abs/1905.02244}.

\bibitem[Tan and Le(2019)]{efficientnetv1}
Mingxing Tan and Quoc~V. Le.
\newblock Efficientnet: Rethinking model scaling for convolutional neural
  networks.
\newblock \emph{CoRR}, abs/1905.11946, 2019.
\newblock URL \url{http://arxiv.org/abs/1905.11946}.

\bibitem[Tan and Le(2021)]{efficientnetv2}
Mingxing Tan and Quoc~V. Le.
\newblock Efficientnetv2: Smaller models and faster training.
\newblock \emph{CoRR}, abs/2104.00298, 2021.
\newblock URL \url{https://arxiv.org/abs/2104.00298}.

\bibitem[He et~al.(2015)He, Zhang, Ren, and Sun]{resnetv1}
Kaiming He, Xiangyu Zhang, Shaoqing Ren, and Jian Sun.
\newblock Deep residual learning for image recognition.
\newblock \emph{CoRR}, abs/1512.03385, 2015.
\newblock URL \url{http://arxiv.org/abs/1512.03385}.

\bibitem[He et~al.(2016)He, Zhang, Ren, and Sun]{resnetv2}
Kaiming He, Xiangyu Zhang, Shaoqing Ren, and Jian Sun.
\newblock Identity mappings in deep residual networks.
\newblock \emph{CoRR}, abs/1603.05027, 2016.
\newblock URL \url{http://arxiv.org/abs/1603.05027}.

\bibitem[Simonyan and Zisserman(2014)]{vgg}
Karen Simonyan and Andrew Zisserman.
\newblock {Very deep convolutional networks for large-scale image recognition}.
\newblock \emph{arXiv preprint arXiv:1409.1556}, 2014.

\bibitem[Szegedy et~al.(2015)Szegedy, Vanhoucke, Ioffe, Shlens, and
  Wojna]{inceptionv3}
Christian Szegedy, Vincent Vanhoucke, Sergey Ioffe, Jonathon Shlens, and
  Zbigniew Wojna.
\newblock Rethinking the inception architecture for computer vision.
\newblock \emph{CoRR}, abs/1512.00567, 2015.
\newblock URL \url{http://arxiv.org/abs/1512.00567}.

\bibitem[Deng et~al.(2019)Deng, Zhao, Yin, Dustdar, and Zomaya]{ai-on-edge}
Shuiguang Deng, Hailiang Zhao, Jianwei Yin, Schahram Dustdar, and Albert~Y.
  Zomaya.
\newblock Edge intelligence: the confluence of edge computing and artificial
  intelligence.
\newblock \emph{CoRR}, abs/1909.00560, 2019.
\newblock URL \url{http://arxiv.org/abs/1909.00560}.

\bibitem[Google(2020{\natexlab{a}})]{coral}
Google.
\newblock Google coral ai accelerator datasheet.
\newblock \url{https://coral.ai/docs/}, 2020{\natexlab{a}}.

\bibitem[Google(2023{\natexlab{b}})]{edge-tpu}
Google.
\newblock Google edge tpu.
\newblock \url{https://cloud.google.com/edge-tpu}, 2023{\natexlab{b}}.

\bibitem[NVIDIA(2023{\natexlab{a}})]{jetson-nano}
NVIDIA.
\newblock Jetson modules.
\newblock \url{https://developer.nvidia.com/embedded/jetson-modules},
  2023{\natexlab{a}}.

\bibitem[NVIDIA(2023{\natexlab{b}})]{jetpack}
NVIDIA.
\newblock Jetpack sdk.
\newblock \url{https://developer.nvidia.com/embedded/jetpack},
  2023{\natexlab{b}}.

\bibitem[NVIDIA(2023{\natexlab{c}})]{jetson-use-cases}
NVIDIA.
\newblock Embedded systems for product development.
\newblock
  \url{https://www.nvidia.com/en-us/autonomous-machines/embedded-systems/product-development/},
  2023{\natexlab{c}}.

\bibitem[Intel(2019)]{ncs2}
Intel.
\newblock Neural compute stick 2 product brief.
\newblock
  \url{https://www.intel.com/content/dam/support/us/en/documents/boardsandkits/neural-compute-sticks/NCS2_Product-Brief-English.pdf},
  2019.

\bibitem[Deng et~al.(2009)Deng, Dong, Socher, Li, Li, and Fei-Fei]{imagenet}
Jia Deng, Wei Dong, Richard Socher, Li-Jia Li, Kai Li, and Li~Fei-Fei.
\newblock Imagenet: A large-scale hierarchical image database.
\newblock In \emph{2009 IEEE Conference on Computer Vision and Pattern
  Recognition}, pages 248--255, 2009.
\newblock \doi{10.1109/CVPR.2009.5206848}.

\bibitem[Dosovitskiy et~al.(2020)Dosovitskiy, Beyer, Kolesnikov, Weissenborn,
  Zhai, Unterthiner, Dehghani, Minderer, Heigold, Gelly, Uszkoreit, and
  Houlsby]{vit}
Alexey Dosovitskiy, Lucas Beyer, Alexander Kolesnikov, Dirk Weissenborn,
  Xiaohua Zhai, Thomas Unterthiner, Mostafa Dehghani, Matthias Minderer, Georg
  Heigold, Sylvain Gelly, Jakob Uszkoreit, and Neil Houlsby.
\newblock An image is worth 16x16 words: Transformers for image recognition at
  scale.
\newblock \emph{CoRR}, abs/2010.11929, 2020.
\newblock URL \url{https://arxiv.org/abs/2010.11929}.

\bibitem[Google(2020{\natexlab{b}})]{pycoral-repo}
Google.
\newblock Pycoral api repository.
\newblock \url{https://github.com/google-coral/pycoral}, 2020{\natexlab{b}}.

\bibitem[Foundation(2023)]{timeit}
Python~Soft. Foundation.
\newblock Timeit package api.
\newblock \url{https://docs.python.org/3/library/timeit.html}, 2023.

\bibitem[ONNX(2021)]{tf2onnx}
ONNX.
\newblock tf2onnx repository.
\newblock \url{https://github.com/onnx/tensorflow-onnx}, 2021.

\bibitem[Intel(2022{\natexlab{a}})]{openvino-dev}
Intel.
\newblock Openvino-dev download page.
\newblock
  \url{https://www.intel.com/content/www/us/en/developer/tools/openvino-toolkit/download.html},
  2022{\natexlab{a}}.

\bibitem[Intel(2022{\natexlab{b}})]{openvino-package}
Intel.
\newblock Openvino toolkit packages.
\newblock
  \url{https://storage.openvinotoolkit.org/repositories/openvino/packages/2022.2/linux},
  2022{\natexlab{b}}.

\bibitem[NVIDIA(2023{\natexlab{d}})]{tensorrt}
NVIDIA.
\newblock Tensorrt.
\newblock \url{https://developer.nvidia.com/tensorrt}, 2023{\natexlab{d}}.

\end{thebibliography}






\end{document}